\documentclass{aastex631}

\usepackage{graphicx}	
\usepackage{amsmath}	
\usepackage{amssymb}	
\usepackage{natbib}

\shorttitle{Moving Objects in Earth-Observation images}
\shortauthors{Keto et al.}
\graphicspath{{./}{figures/}}

\begin{document}

\title{Detection of Moving Objects in Earth Observation Satellite Images}

\author{Eric Keto}
\affiliation{Harvard University, Institute for Theory and Computation \\
60 Garden Street,
Cambridge, MA USA}

\author{Wesley Andr\'{e}s Watters}
\affiliation{Wellesley College, Whitin Observatory \\
106 Central St.,
Wellesley, MA 02481, USA}


\begin{abstract}
Moving objects have characteristic signatures in multi-spectral images made by Earth observation satellites that use push broom scanning. 
While the general concept is applicable to all satellites of this type, each satellite design has its own unique imaging system and requires 
unique methods to analyze the characteristic signatures.
We assess the feasibility of detecting moving objects and measuring
their velocities in one particular archive of satellite images made by Planet Labs Corporation 
with their constellation of SuperDove 
satellites. Planet Labs data presents a particular challenge in that the images in the archive are mosaics of individual exposures 
and therefore do not have
unique time stamps. We explain how the timing information can be restored indirectly. Our results indicate that the movement of common
transportation vehicles, airplanes, cars, and boats, can be detected and measured.

\end{abstract}

\keywords{METHODS: DATA ANALYSIS, TECHNIQUES: IMAGE PROCESSING}

\section{Introduction} \label{introduction}

Satellites that survey large areas of the earth generally collect single-frame images
rather than full-motion video. Nonetheless, the motion of observed objects
creates characteristic signatures in single-frame images obtained by push broom  scanning. 
In this technique, if the regions on the camera's sensor that are 
responsible for the different spectral bands are arranged as stripes perpendicular to 
orbital track, then the spectral images are observed at slightly different times 
as the orbital velocity pushes the camera
over the area imaged.
The spectral images represent a video with a frame rate equal to the transit
time and a duration set by the number of spectral bands. If the time scale of the motion of an imaged object is commensurate
with the transit time of the satellite, the moving object appears at a different
location in the images in different spectral bands. In 
the difference of the images in two spectral bands, a moving object appears as a positive and negative pair.
A second signature of motion results from the different colors
of the spectral band images. In a 
full-color composite, the moving object appears as displaced ghosts
in different colors. Both effects depend
on the velocity of the moving object with respect the velocity of the satellite
and also on the characteristics of the imaging system.

Here we explore the feasibility of identifying and characterizing moving objects in one archive of Earth observation images
made by Planet Labs Corporation 
with their latest constellation of satellites named SuperDove \citep{PL22}. 
These satellites image
in seven spectral bands in the optical and one in the near infrared. 
Our results show that objects moving with speeds of typical transportation vehicles, cars, boats,
and aircraft, are easily detectable by eye. The minimum ($\sim 20$ ms$^{-1}$) 
detectable ground speeds is set
by the spatial resolution. There is an ambiguity between
the speed and altitude of an object that is discussed further in section \S\ref{altitude-speed}.

Other studies have exploited timing differences in satellite images to detect motion. The procedures used are specific
to each satellite. For example, the Quickbird satellite imaged the same scene in black and white and color
with an 0.2 s delay that allows detection of motion \citep{Easson2010}. \citet{Heiselberg2019, Heiselberg2021} 
used the same general technique adapted to images
from the multispectral Sentinal-2 satellite. Here we restrict our study to the data taken by Planet Labs
SuperDove satellites although our discussion is applicable more generally to satellites with similar imaging characteristics.

\section{Observations} \label{observations}

\begin{figure} 
\includegraphics[width=7.0in]{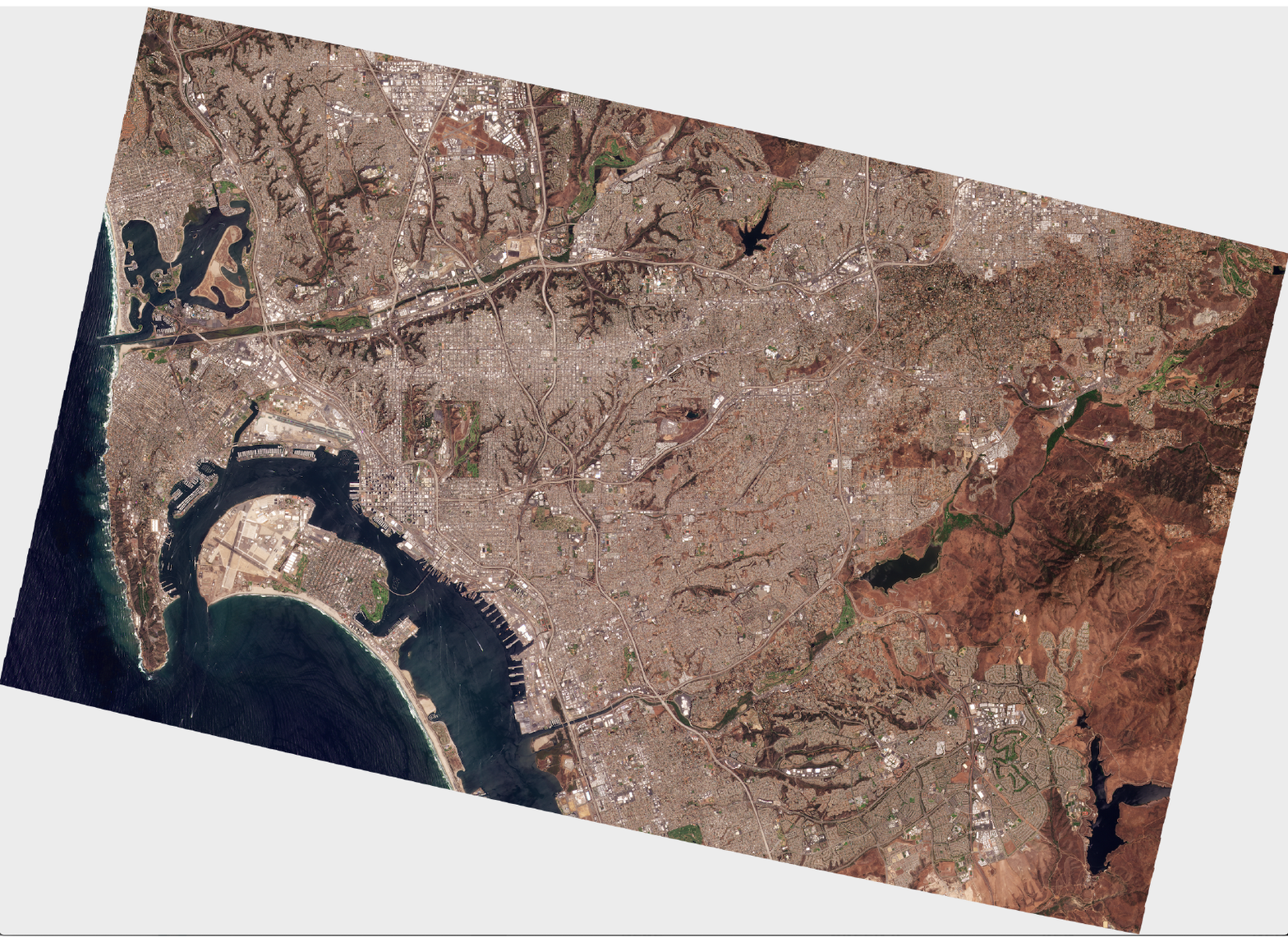}
\caption{
Composite visual image from the Planet Labs data set 20220530\_173806\_19\_241e.
The image contains $13317 \times 9578 $ pixels of $3\time 3$ m$^2$ 
covering $40.0 \times 29.7$ km$^2$ including the regions with no data.
}
\label{visual-image}
\end{figure}

We analyze the images in one scene from the Planet Labs archive with identifier 
20220530\_173806\_19\_241e indicating that the observations were made on May 30, 2022 at 17:38:06 UT by the satellite identified as 241e.  
Each scene in the archive is a data set containing images in each of eight spectral bands as well as a visual composite in RGB (red-green-blue) format.
The images were made with a pixel size on the ground (GSD or ground sample distance) of 4.2 m and covering
$\sim 37 \times 22 \ {\rm km}^2$. The scene includes images processed or orthorectified by Planet Labs
to a longitude and latitude grid and resampled to 3.0 m. 
The orthorectified visual image of the area around San Diego, CA is shown in figure \ref{visual-image}.
 
\section{Characteristic signatures of moving objects}

\subsection{Signature in composite visual images}
\begin{figure}
\includegraphics[width=7.0in]{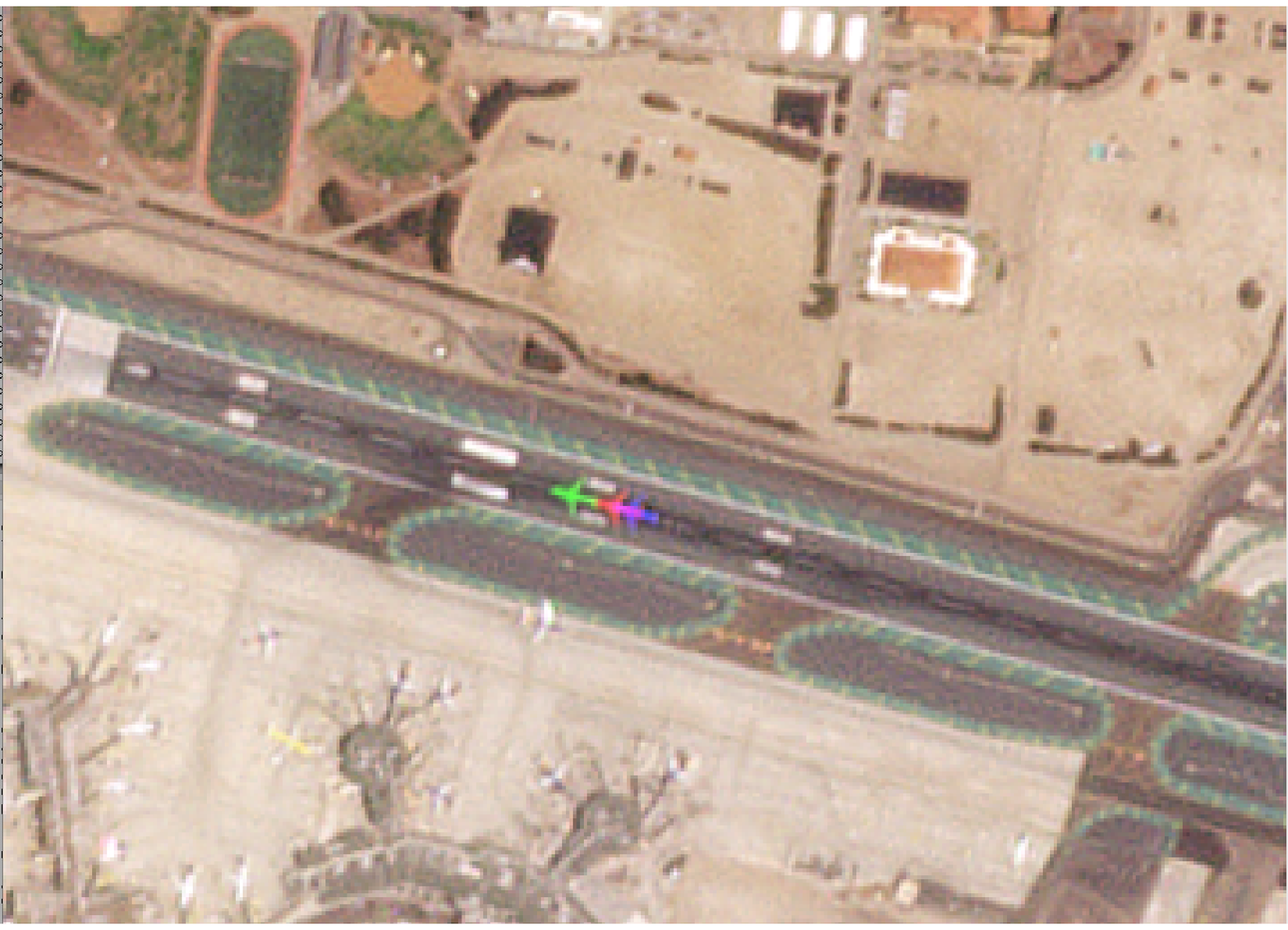}
\caption{
Detail of the composite visual image (figure \ref{visual-image}) showing an airplane moving on or over the
runway at the San Diego International airport. 
The image covers $ 1.13 \times 0.77$  km.}
\label{visual-detail}
\end{figure}

\begin{figure}
\includegraphics[width=7.0in]{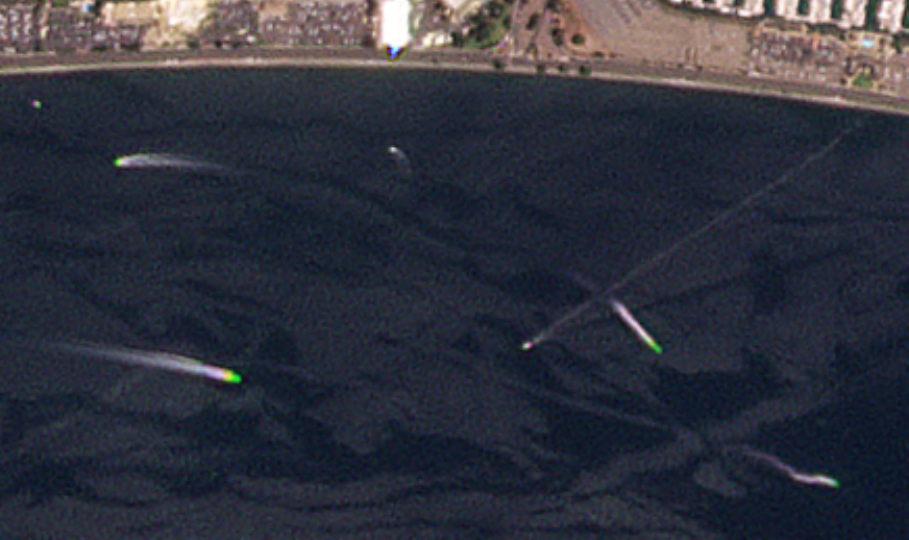}
\caption{
Detail of the composite visual image (figure \ref{visual-image}) showing boats moving in the San Diego
harbor. The bright green leading edge is a characteristic signature of motion.
The image covers $ 1.31 \times 0.78$  km.}
\label{boats}
\end{figure}

A detail of the visual image (figure \ref{visual-detail}) shows an airplane moving on or over the
runway at the San Diego International airport. The motion is
identified by the appearance of three airplanes each in one of the RGB colors.
The separation between these three colored versions of the moving object depends on the time delay in the acquisition
of the images and the velocity of the moving object.  Objects moving too slowly to appear as separate
images, such as the boats in figure \ref{boats},  show
a green leading edge, a blue trailing edge, with red in the middle.


\subsection{Signature in differenced images}
\begin{figure}
\includegraphics[width=7.0in]{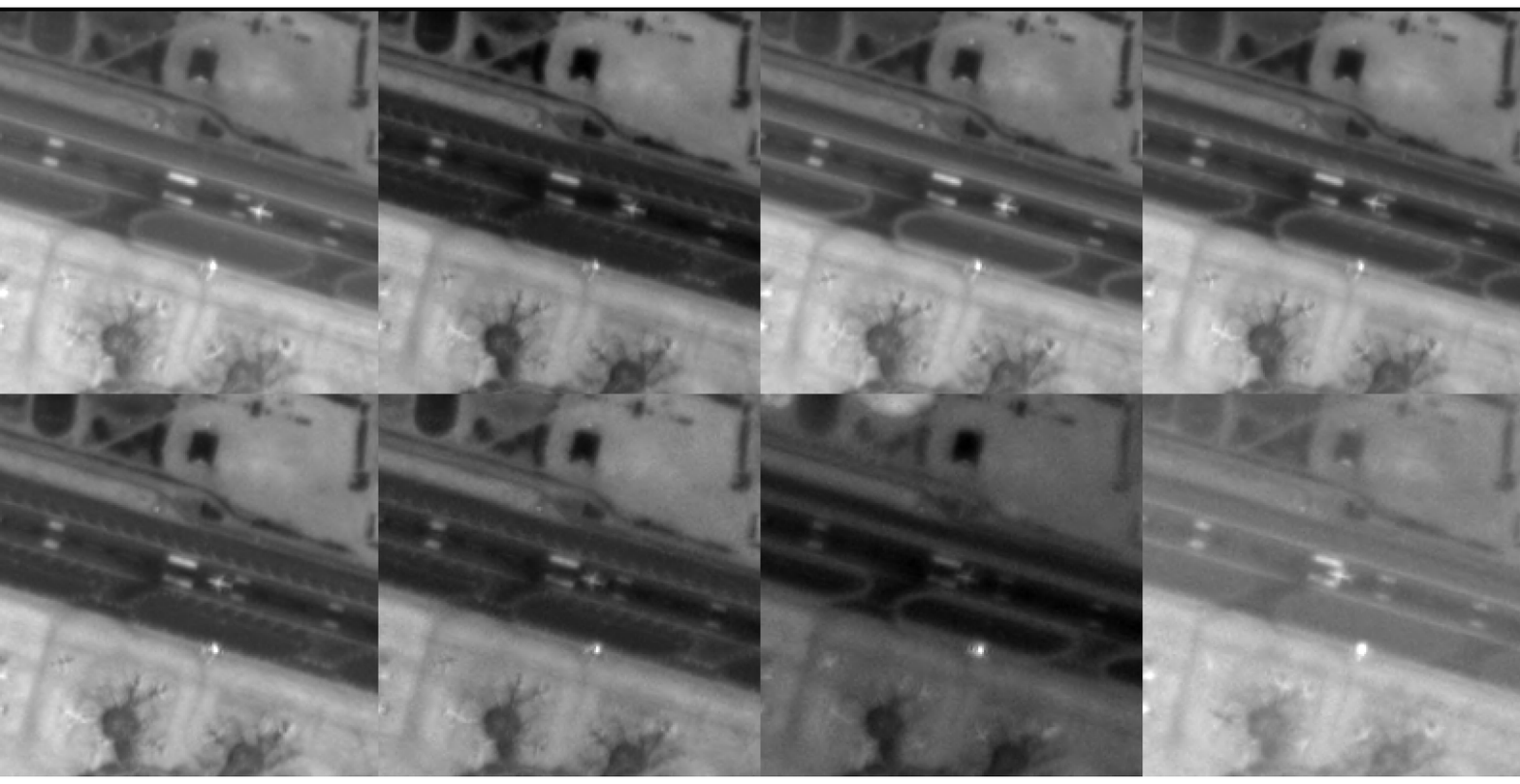}
\caption{
Detail of the full image in eight spectral bands  arranged in temporal order: top row, right-to-left, then bottom row, right-to-left.
In this temporal order showing the airplane moving west (left), the spectral bands are B, R, G1, G2, Y, RE, NIR, CB (table \ref{T2}).
Each panel covers  $0.6 \times 0.6$ km.}
\label{8-band}
\end{figure}

\begin{figure}
\includegraphics[width=7.0in]{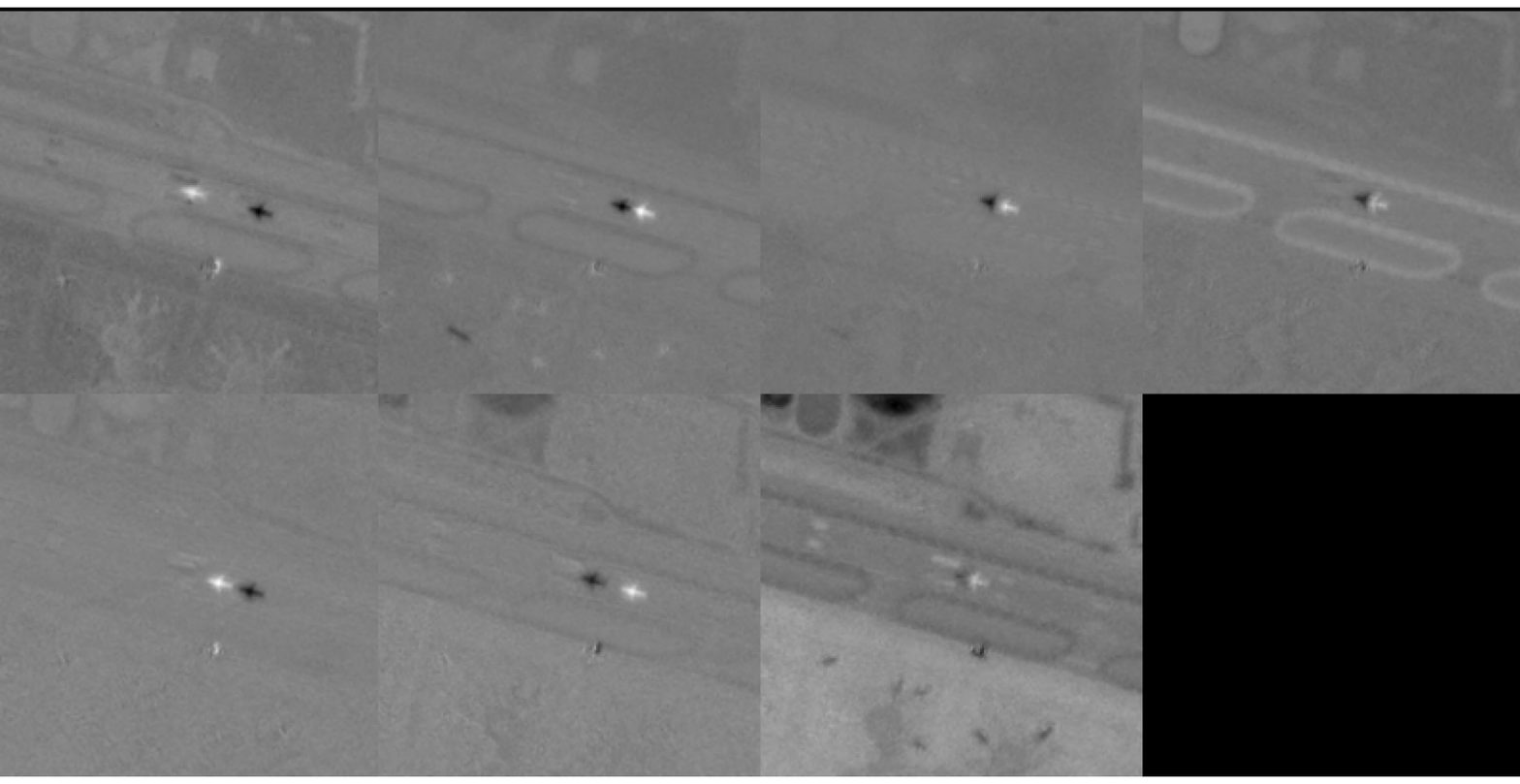}
\caption{
Differences of the images in figure \ref{8-band} to isolate the moving object. The background is flatter if the images are subtracted
in spectral order. Thus the 7 images show the differences CB-B, B-G1, G1-G2, G2-Y, Y-R, R-RE, RE-NIR (table \ref{T2}). 
The eighth square (black) has no data.
}
\label{8-band-difference}
\end{figure}

\begin{figure}
\includegraphics[width=7.0in]{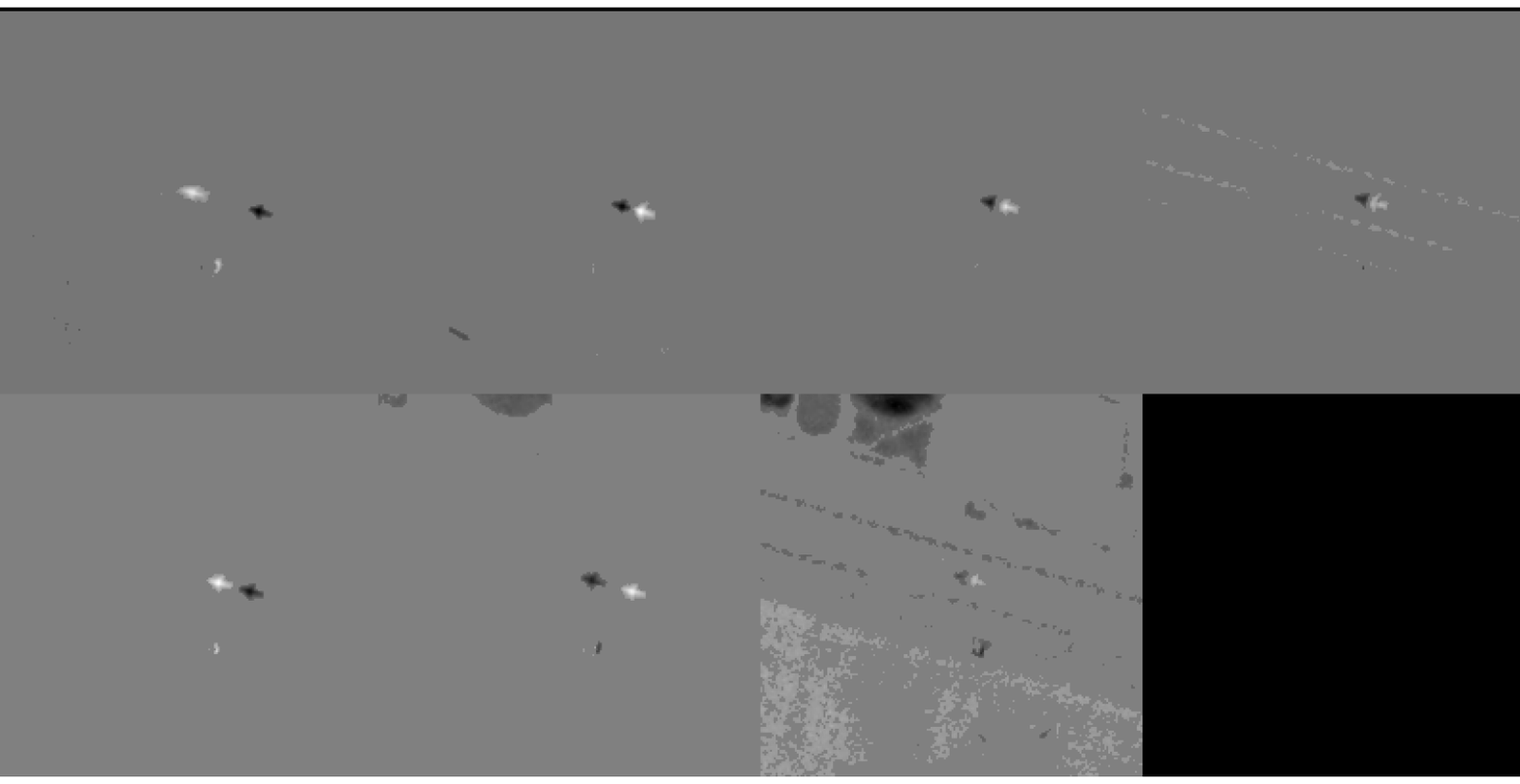}
\caption{
Same as figure \ref{8-band-difference} after thresholding the images at a constant level based on the histogram of
image brightness levels. The same threshold has been applied to all images.
}
\label{8-band-threshold}
\end{figure}

Figure \ref{8-band} shows the airplane
of figure \ref{visual-detail} in the eight spectral bands of the SuperDove satellite arranged sequentially in
time to show the airplane moving from right to left along the runway. 
In a differenced image, a moving object appears as
a positive and negative pair highlighted against a suppressed background (figure \ref{8-band-difference}). 
To improve the suppression of the background, the differences are of 
bands that are adjacent in the spectrum rather than sequential in time. To further
suppress the background, the pixels with brightness levels below a 5\% threshold are set
to zero (figure \ref{8-band-threshold}). Difference imaging is commonly used to detect motion 
in remote sensing, surveillance, computer vision, and astronomy and sophisticated techniques
have been developed for different image characteristics \citep{Alard2000, Bramich2013}. 
Our differenced images show that a very simple technique is adequate
with the different spectral bands of the SuperDove satellite.

\section{Millisecond timing from static images}

Planet Labs aims to provide images of the entire land surface of the Earth once per day.
The relevant time scale for the acquisition of a single image is on the order of minutes, sufficient to correlate with
the illumination angle of the sun and the weather, for example. Furthermore, the images in the archive
are mosaics of several camera exposures or frames that were taken at different times as the satellite crossed a region on the ground.
The images in the archive do not have exact time stamps but rather the approximate time over which the
several images in a mosaic were obtained.
In order to measure the velocity of moving objects by comparison of images in different spectral bands,
we require the difference in the acquisition time with an accuracy
of 1 - 10 ms. This requires an understanding of the imaging process.

The sensor on the SuperDove satellite has 8880 x 5304 pixels each with a size of $ 5.5 \times 5.5$ $\mu$m$^2$. In a coordinate
system oriented with the orbital track, we label the axes $x$ and $y$ with the satellite moving along the y-axis
consistent with the presentation of the ``basic analytic" images in the Planet Labs archive. 
The sensor is covered by a filter with 8 regions of 8880 x 663 pixels, one for each of
the 8 spectral bands. Each individual camera exposure captures these 8 color strips, 
each covering
a different region on the ground. 

After a time delay, a second image is taken. 
This second image is shifted with respect to the first along the ground track of the satellite
by the orbital motion. The time delay, set by the maximum frame rate of the camera ($\sim 0.17$ s),
is shorter than the time required for one color to exactly overlap another on the ground. Repeated, the process
results in a continuous mosaic of partially overlapping mages in each color. Where there is overlap,
the image-processing pipeline selects one row
from the rows of overlapping pixels. 

\begin{figure}
\includegraphics[width=3.25in]{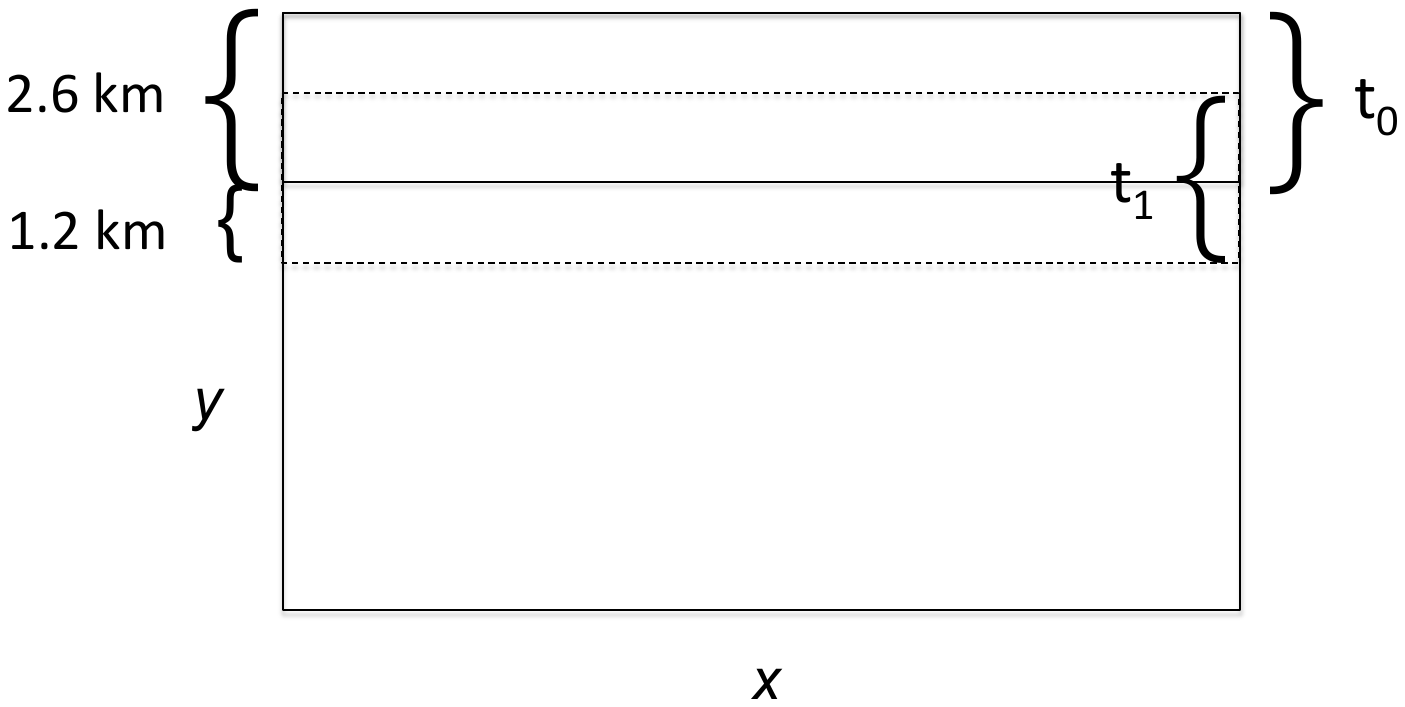}
\caption{
Schematic of the mosaicing process. The large rectangle represents the mosaiced image with a width (dimension in y or vertical)
of 20.8 km. 
The smaller rectangle in solid lines represents the width of one color strip, approximately 2.6 km on the ground, taken at
time $t_0$. After a delay of one interval of the camera frame rate of 0.17 s, the color strip has moved approximately 1.2 km
to the position of the rectangle with dashed horizontal lines. The acquisition time of this strip is $t_1$. Depending on the
mosaicing pattern, a pixel in the region of overlap may have either acquisition time. The top of the color strip $t_0$ might
not line up with the top of the mosaiced image as suggested in the figure, but the procedure suggested in the text
remains valid.
}
\label{mosaic}
\end{figure}

The process is illustrated in figure \ref{mosaic} showing the construction of a mosaic from several strips. 
From the full image acquired at time, $t_0$, extract the color strip for one spectral band to start the mosaic. Repeat for the
image acquired at time, $t_1$, after the orbit of the satellite has shifted the image on the ground. With a frame rate,
$(t_1-t_0)^{-1}$ rapid enough to provide spatial overlap, align the two overlapping strips and select one of the
two overlapping regions. Repeat to fill out the mosaic. Each 
row of pixels in the mosaic then has the acquisition time of the strip that
contributes the row to the mosaic.

Finally, the continuous strip is divided into individual scenes with the same dimensions
as the camera sensor. The basic analytic images for one scene in the archive consist of 
8 images, one in each of the 8 spectral bands,
each with the same number of pixels as the sensor. 
However, each image is a mosaic of several strips of the same color 
captured sequentially as the satellite passes over the 
scene.  The identical process operates for each spectral band.

We break the calculation of the acquisition time difference between pixels in the different spectral bands into two steps. 
In the first step, we find the time difference between the mosaiced images of the eight spectral bands. 
Because the progression of acquisition times across the mosaic is set by the frame rate of the camera
and the orbital speed of the satellite, the progression is the same for all colors. 
This defines a single time difference between two images of different colors even though the image of each color includes
multiple acquisition times. 

This first time difference between spectral bands depends only on the orbital velocity and the sensor dimensions 
and is given by the width of a color strip on the ground divided by the ground speed of the
satellite and the number of color strips on the sensor between the two colors. 
The ground pixel size or ground sample distance (GSD), defined here as the 
size of a pixel on the sensor multiplied by the ratio of the altitude
of the satellite over the focal length of the camera,
is available in the Planet Labs 
archive for the basic analytic images. The orbital speed of the satellite can be determined from the orbital parameters 
published in a daily ephemeris for all Planet Labs satellites mainly for the purpose of collision avoidance with other spacecraft.
For the purposes of this example, we can assume that the orbit is circular and the Earth is spherical. This allows us to
use just one of the published parameters, the mean motion or the number of
orbits per day. In this case, the time difference
between adjacent colors on the sensor is,
\begin{equation}
\Delta t_{color} = \frac {N_y \mu} {2\pi R_\oplus \omega}
\label{eqn1}\end{equation}
where $N_y=663$ is the width in pixels of one color strip, $R_\oplus = 6378$ km is the radius of the Earth, $\mu$ is the GSD of 4 m,
and $\omega$ is the mean motion
of 15.15 orbits per day. The time difference between colors is $\Delta t_{color} \sim 0.39$ s.

The second component of the total time difference  depends on the
location of the object in the mosaiced image of a single color and the progression of acquisition times of the color strips
 that make up the 
mosaic. Without the details of the image processing that constructs the mosaic, in particular which of the strips contributes a row
of pixels to the full mosaic, we can try to determine the second component from the apparent acceleration 
of the moving object itself. As a first estimate,
we use the time difference between colors to calculate the velocity from the position
of the object in all pairs of images in different colors to get a time sequence of velocities. 
If the object remains in a width where
the acquisition time is constant, as would be the case for motion across the satellite track, 
this estimate of the delay time is correct. 

The width of the strip depends on
the camera frame rate, the ground velocity of the satellite, and the GSD.
For example, if the camera frame rate expressed as an interval is 0.17 s, then the ground motion of the image is 1.2 km between
frames, and the pixels within some fraction of this width, depending on the mosaicing pattern, 
have a constant acquisition time.
An object near the boundary of a strip could move into the adjacent strip with a different acquisition time. 
In this case, the time delay 
between the positions of the objects needs to be adjusted by the camera frame interval, $\Delta  t_{camera}$. 
In general, the measured velocity is,
\begin{equation}
v_i = \frac {\Delta p} {\Delta t_{color} + a}
\label{eqn2}\end{equation}
where $\Delta p = p_i(x,y) - p_{i+1}(x,y)$ is the distance between the locations of the object in two color bands, $i$ and $i+1$. 
The variable $a$ can take one of 3 values,  $0, \pm \Delta  t_{camera}$. We can estimate the correct value by minimizing the
apparent acceleration between the measured velocities. Since $\Delta t_{color} \sim 2 \Delta t_{camera}$, an object crossing a boundary
between two acquisition times 
would appear to change velocity by a factor of about 0.5 or 1.5 compared to the other velocities in the time sequence of velocities 
in a time interval equal to the first time difference, $\sim 0.39$ s. For more rapidly moving objects,
this is more likely to be due to the difference in acquisition times than the actual acceleration of the object. 

\section{ Measuring the velocity}

In the first panel in figure \ref{8-band-threshold} showing the thresholded differenced images, 
the airplane is moving approximately perpendicular to the orbital track and therefore
we can use the time delay between colors to estimate the velocity. From equations \ref{eqn1} and \ref{eqn2}
and the data for satellite 241e,
the time delay is 0.393562 s. 
The displacement between the positive and 
negative images of the airplane is 49.6 pixels of 3 m at ground level. 
The average speed of the airplane in the time between these images
is $38.5 \pm 5.3$ m/s or $86.4 \pm 11.8$ mph. For reference, the average speed of a Boeing 737 at takeoff is 150 mph.

The error in the measured velocity is dominated by the pixelization and also affected by the ability to define
a moving object against a complex background. For objects easily distinguished above the background, the
error in location should be a pixel or less. The expected error in the velocity is then about 11 ms$^{-1}$. The
expected error in acceleration from 3 positions would be $\sqrt{3/2}$ larger. We can ignore the
error contributed by the smaller uncertainty in time. Since the time delay is proportional to square root of
the orbital altitude, ($\sqrt{r}$), the error in the time delay is $\delta r/ (2\sqrt{r})$. The expected error in
the altitude of a satellite in low earth orbit, $\delta r$, described by a two-line element (TLE) model is about 0.1 km \citep{Floher2008}.
Thus the fractional error in time is on the order of 0.002.


\subsection{Altitude-speed ambiguity}\label{altitude-speed}

\begin{figure}
\includegraphics[width=3.25in]{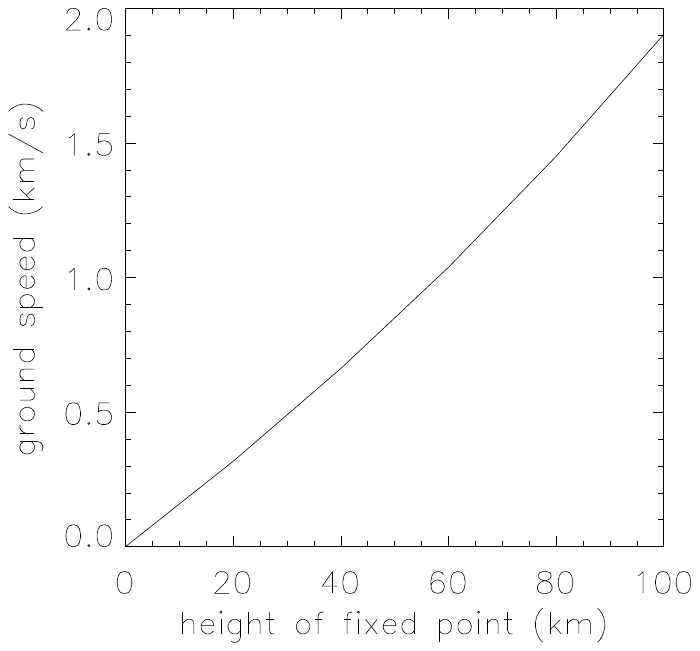}
\caption{
Apparent ground speed (vertical axis) versus altitude (horizontal axis) of an object that is
stationary with respect to a location on the Earth's surface.}
\label{a-s}
\end{figure}

The spatial size of the pixels in the image (GSD) depends linearly on the orbital altitude, $\sim 500$ km,
and the elevation of the Earth's surface. 
If the moving object is itself at some altitude above the ground, then the actual
distance traveled by the object between images and its velocity 
will be less than indicated by the number of pixels and the GSD. 
Since top of the stratosphere is $\sim 50$ km, the ambiguity results in an uncertainty of less than 10\% for objects
flying in the atmosphere. Of course, there may be objects moving outside the atmosphere whose speed 
may be very different from the speed calculated at ground level. 
Alternatively, stationary objects at altitude may show misaligned colors similar to the signature of motion
but due to parallax
caused by the spatial alignment which was referenced to the ground in order to eliminate parallax at the ground level.
For example, figure \ref{a-s} shows the apparent ground speed that would be derived for
an object that was at altitude and stationary with respect to the Earth's surface. 
\begin{equation}
v_{app} = \frac {  h_{obj}  }    {  h_{sat} - h_{obj}  } \frac{L}{\Delta t}
\end{equation}
where $h_{obj}$ and $h_{sat}$ are the respective altitudes of the object and satellite, and $L$ is the linear 
distance
traveled by the satellite in time $\Delta t$ between images.
Clouds at altitude are commonly seen with a signature of motion in Planet Labs images.

In some cases, the altitude-speed ambiguity can be resolved from information in the image itself. For example,
if the direction of an object's motion can be determined, and the object is not moving parallel to
the satellite's orbit, the object's apparent motion can be resolved into two components: the actual
velocity of the object, and an apparent velocity due to the parallax created by the satellite motion
\citep{Heiselberg2021}. The latter allows a determination of the object's
altitude. However, the direction of motion may not be obvious. For example,
crosswinds may cause the orientation of an airplane to differ from its direction of motion.

\section{Conclusions and Future research}\label{future}

This research demonstrates at least two characteristic signatures of motion in Earth observation images made with
push broom scanning and demonstrates a method for estimating the velocity. 
The two signatures, different-colored ghosts in visual images and positive-negative pairs in
the differenced images of individual spectral bands, are distinctive and easy to recognize by eye. We intend to develop
software that makes use of pattern-recognition techniques to automatically detect any moving objects in Planet Labs images. 

This research is conducted as part of the Galileo Project \citep{Loeb2022} at Harvard University 
 whose goal is to collect
scientific quality data that may be useful in the search for objects of extraterrestrial origin
(https://projects.iq.harvard.edu/galileo/home). For objects within the Earth's atmosphere, such as those that may be detected
in the Planet Labs images, unusual flight patterns
have been suggested as an indicator of extraterrestrial origin \citep{ODNI2021}. This research is a first step to
develop a method to observe large areas of the Earth, automatically recognize moving objects, and select those
whose velocity, acceleration, size or shape fall outside those expected for natural phenomena or common vehicles or 
projectiles \citep{Watters2022}. Satellite data may also be used with other data in this goal. 
For example, the Galileo Project is building a network of
ground based observatories to collect visual and infrared images of flying objects at much higher spatial resolution
than available in satellite images \citep{Szenher2022}. Historical aircraft transponder broadcasts of heading, speed, 
and altitude are also
available to identify common terrestrial aircraft. Historical meteorological data can help identify natural 
atmospheric phenomena.

\begin{deluxetable*}{lccc}
\tablecaption{Spectral Bands of SuperDove Satellites}
\tablewidth{0pt}
\tablehead {
\colhead {Color name} & \colhead{Short name} & \colhead{Coverage}   & Position\\
                                    &                                     &\colhead {(nm)  } 
}
\decimalcolnumbers
\startdata
Coastal Blue 	& CB & 431 - 452  & 7\\
Blue			& B    & 465 - 515 & 0 \\
Green I		& G1  & 513 - 549 & 2\\
Green II		& G2  & 547 - 583 & 3\\
Yellow		& Y    & 600 - 620 & 4\\
Red			& R    &650 - 680 & 1\\
Red-Edge		& RE  &697 - 713 & 5\\
NIR			& NIR &845 - 885 & 6\\
\enddata
\tablecomments{The position refers to the location of the spectral filter with respect to the 
direction of the satellite. The Blue filter is over the first 663 pixels of the sensor with respect to
the satellite motion. Thus the Blue band observes a section of ground ahead of the other
bands. Similarly Coastal Blue observes a section of ground behind the other bands. After 8
exposures, a section of ground is observed in all bands.}
\label{T2}
\end{deluxetable*}

\begin{acknowledgments} 
We thank Planet Labs for access to data and technical support.
\end{acknowledgments}

\bibliography{sat}{}

\end{document}